\documentclass[letterpaper, 10 pt, conference]{ieeeconf}

\IEEEoverridecommandlockouts                              %

\usepackage{graphicx} %
\usepackage{subcaption}
\usepackage{tikz}
\usepackage{hyperref}

\usepackage{multirow}
\usepackage{boldline}
\usepackage{multicol, booktabs,}
\usepackage{wrapfig}
\usepackage{times} %
\usepackage{amsmath} %

\title{Belief Regulated Dual Propagation Nets for Learning Action  Effects on Groups of Articulated Objects}

\author{Ahmet E. Tekden$^{1}$, Aykut Erdem$^{2}$, Erkut Erdem$^{2}$, Mert Imre$^1$, M. Yunus Seker$^1$ and Emre Ugur$^1$%
\thanks{*This research has received funding from the European Union's Horizon 2020 research and innovation programme under grant agreement no. 731761, IMAGINE; supported by a TUBA GEBIP fellowship awarded to E. Erdem; and supported by a Tubitak 2210-A scholarship awarded to A.E. Tekden}%
\thanks{$^{1}$Ahmet E. Tekden, Mert Imre, M. Yunus Seker and Emre Ugur are with Computer Engineering Department, Bogazici University, Turkey.}%
\thanks{$^{2}$Aykut Erdem and Erkut Erdem are with Computer Engineering Department, Hacettepe University, Turkey}
\thanks{*Corresponding author: Ahmet E. Tekden, email: \url{ercan.tekden@boun.edu.tr} }
}
\usepackage{xcolor}
\definecolor{darkgreen}{rgb}{0.0, 0.5, 0.0}

\begin{document}

\maketitle
\thispagestyle{empty}
\pagestyle{empty}

\begin{abstract}
In recent years, graph neural networks have been successfully applied for learning the dynamics of complex and partially observable physical systems. However, their use in the robotics domain is, to date, still limited. In this paper, we introduce \emph{Belief Regulated Dual Propagation Networks (BRDPN)}, a general-purpose learnable physics engine, which enables a robot to predict the effects of its actions in scenes containing groups of articulated multi-part objects. Specifically, our framework extends recently proposed propagation networks (PropNets) and consists of two complementary components, a \emph{physics predictor} and a \emph{belief regulator}. While the former predicts the future states of the object(s) manipulated by the robot, the latter constantly corrects the robot’s knowledge regarding the objects and their relations.  Our results showed that after training in a simulator, the robot can reliably predict the consequences of its actions in object trajectory level and exploit its own interaction experience to correct its belief about the state of the environment,
enabling better predictions in partially observable environments. Furthermore, the trained model was transferred to the real world and verified in predicting trajectories of pushed interacting objects whose joint relations were initially unknown. We compared BRDPN against PropNets, and showed that BRDPN performs consistently well. Moreover, BRDPN can adapt its physic predictions, since the relations can be predicted online. 
\end{abstract}

\section{INTRODUCTION}

In complex robotic systems, predicting the effects is a challenging problem when the number of objects varies, especially in the presence of rich and various interactions among these objects. Previously, we investigated predicting low-level effects on arbitrary shaped single objects given robot actions  \cite{SEKER2019173}. On the other hand, when several objects are linked with physical connections, this would also suggest some semantic connections between them, such as the motion of one object can propagate its motion onto another object, which might lead to a chain effect. To be able to model such systems accurately, the representation of data should be able to encode multiple objects and their interactions with each other, and it should be robust to perturbations. Recently, a great amount of effort has put on the prediction of the dynamics via graph networks (e.g. ~\cite{battaglia2016interaction,chang2016compositional,li2018propagation,mrowca2018flexible,li2018learning,watters2017visual,van2018relational}). These works can deal with varying number of objects and learn rich interaction dynamics among these objects. Some of these works have focused on unsupervised learning, while others were aimed at developing learnable physics engines. However, applying them to model robot-object interactions is not very straightforward as the active involvement of the robot was not taken into account and, uncertainty in perception was not explicitly addressed.

 In this work, we propose \emph{Belief Regulated Dual Propagation Network (BRDPN)} which takes the actions of the robot and their effects
into account in predicting the next states\footnote{Our source code and experimental data are available in the web page prepared for this paper: \url{https://fzaero.github.io/BRDPN/}.}. 
The network continuously regulates its belief about the environment based on its interaction history to correct its future predictions. For belief regulation, extending the recently proposed propagation networks (PropNets)~\cite{li2018propagation} that handle instantaneous effect propagation, we propose a temporal propagation network that takes history of the motion of each object to predict unknown object or relation properties. Our system is verified on a table-top push setup that has cylindrical objects and joint relations between them. Our setup includes varying number of objects that might be connected with \emph{rigid}, \emph{revolute} or \emph{prismatic joints}. The model definitions of these types of relations, including the PropNet$_{n}$ relation, is not provided to the robot. 
Notice that in our settings the relations between objects cannot be perceived by the robot. From its interaction experience in the simulator, it learns to predict relations between objects given observed object motions, and exploits this information to predict future object trajectories. 
Furthermore, it was transferred to real world and verified in experiments that included around 100 interactions. Our system was shown to outperform the original PropNets, both in simulation and real-world, when the relations between objects were not reliably provided to the system.

Our contribution to the state of the art is two-fold. First, we introduced a deep neural network based  method for learning how to exploit the interaction experience of the robot to extract values of otherwise unknown state variables in partially observable environments. Second, we implemented a learning based effect prediction robotic framework that can handle multiple interacting objects that might have different types of connections, and we verified this framework both in simulated and real robot experiments.

\section{Related Work}
There is a considerable progress in modelling physics with probabilistic approaches in recent years. For instance, Battaglia et al. \cite{battaglia2013simulation} proposed a Bayesian model called Intuitive Physics Engine and showed that the physics of stacked cuboids can be modelled with this model. Deisenroth et al. \cite{deisenroth2011pilco} suggested a probabilistic dynamics model that depends on Gaussian Processes,and is capable of predicting the next state of a robot given the current state and its actions. Recently, some researchers extended these works by using deep learning methods to model physics. Wu et al. \cite{wu2015galileo} proposed a deep approach for finding the parameters of a simulation engine which predicts the future positions of the objects that slide on various tilted surfaces. Lerer et al.~\cite{lerer2016learning} trained a deep network to predict the stability of the block towers given their raw images obtained from a simulator. A specific topic of interest within modelling physics with deep learning is motion prediction from images, which has gained increasing attention over the last few years. The studies presented for this task either employ convolutional neural network (CNNs) or graph neural networks (GNNs).

Mottaghi et al. \cite{mottaghi2016newtonian} trained a CNN for motion prediction on static images by casting this problem as a classification problem. Mottaghi et al. \cite{mottaghi2016happens} employed CNNs to predict movements of objects in a static image when some external forces are applied to them. Fragkiadaki et al. \cite{fragkiadaki2015learning} suggested a deep architecture in which the outputs of a CNN are used as inputs to Long Short Term Memory (LSTM) cells~\cite{hochreiter1997long} to predict movements of balls in simulated environments. 

A number of studies have examined the action-effect prediction in videos. Finn et al.~\cite{finn2016unsupervised} proposed a convolutional recurrent neural network~\cite{xingjian2015convolutional} to predict the future image frames using only the current image frame and actions of the robot. Byravan et al. \cite{byravan2017se3} presented an encoder-decoder like architecture to predict SE(3) motions of rigid bodies in-depth data. However, the output images get blurry over time or their predictions tend to drift away from the actual data due to the accumulated errors, making it not straightforward to use for long-term predictions in robotics.

As deep structured models, GNNs allow learning useful representations of entities and relations among them, providing a reasoning tool for solving structured learning problems. Hence, it has found particularly wide use in physics prediction. Interaction network by Battaglia et al.~\cite{battaglia2016interaction} and Neural Physics Engine by Chang et al. \cite{chang2016compositional} are the earliest examples to general-purpose physic engines that depend on GNNs. These models do object-centric and relation-centric reasoning to predict movements of objects in a scene. Though they were successful in modelling dynamics of several systems such as n-body simulation and billiard balls, their models had certain shortcomings, especially when an object's movement has chain effects on other objects (\textit{e.g.} a pushed object pushes other object(s) it is contacting with) or when the objects in motion have complex shapes. These shortcomings can be partly handled by including a message passing structure within GNNs as done in the recent works such as~\cite{li2018propagation,mrowca2018flexible,li2018learning}. Watters et al. \cite{watters2017visual} and van Steenkiste et al. \cite{van2018relational} proposed hybrid network models which encode object information directly from images via CNNs and which predict the next states of the objects with the use of GNNs. Our framework differs from these GNN approaches in that it learns to predict the relations between objects from the interaction history, and uses these relation predictions in estimating the future states of the objects. Additionally, different from these approaches, our model is verified in a real robotic setup.

\section{Proposed Model}
\begin{figure*}[t]
    \centering
    \includegraphics[width=\linewidth]{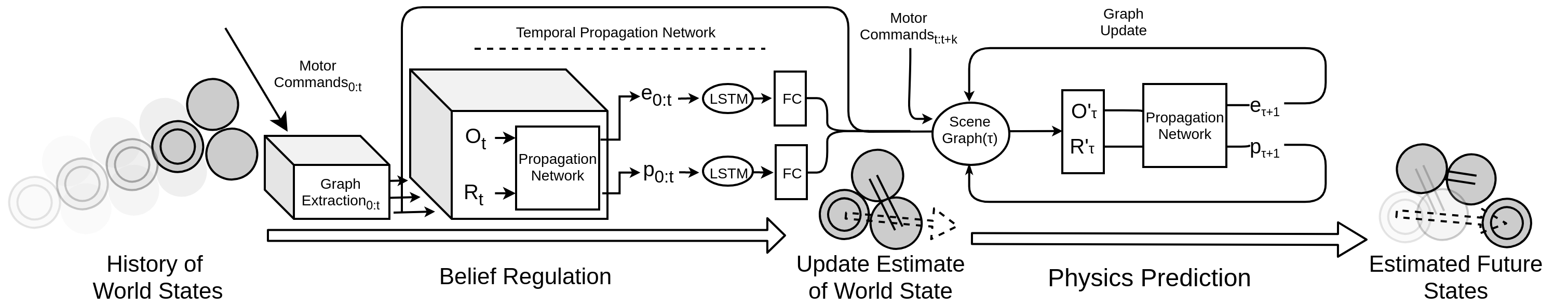}
    \caption{\emph{Belief Regulated Dual Propagation Networks}. The system contains two parts: belief regulation module, and physics prediction module. Given previous world state values and motor commands, the belief regulation module is used to update the estimate of the state variables. 
    Given the current estimate of the world state and planned motor commands, the physics prediction module predicts the sequence of future states expected to be observed.
    }
    \label{fig:complete_system}\vspace{-0.3cm}
\end{figure*}

In this section, we introduce the \emph{Belief Regulated Dual Propagation Networks (BRDPN)} and explain how it extends the propagation network framework for articulated multi-part multi-object settings to allow the regulation of the beliefs about environment state variables. Belief regulation corresponds to regulating the robot's belief about the environment through extracting or updating the values of state variables. Fig.~\ref{fig:complete_system} shows a graphical illustration of our framework, which is composed of two main components: a \emph{physics predictor} and a \emph{belief regulator}. The physics predictor is based on propagation network and responsible for predicting future states of the manipulated objects. The belief regulation module is a propagation network with recurrent connections, which we call temporal propagation network. Belief regulation module is responsible from extracting/updating the knowledge of the robot about the environment through its observations of own-interaction experience. We explain the technical details of these models in the following parts. 
\paragraph*{Preliminaries} Assume that the robot is operating in a complex environment involving a set of multi-part objects $O$, we express the scene with a graph structure $G=\left\langle O, R\right\rangle$ where the nodes $O = \{o_i\}_{i=1:N^o}$ represent the set of objects (of cardinality $N^o$) and the edges $R = \{r_{k}\}_{k=1:N^r}$ represent the set of relations between them (of cardinality $N^r$). More formally, each node $o_i= \left\langle {x_i},{a^o_i}\right\rangle$ stores object related information, where ${x_i} = \left\langle {q_i},{\dot q_i}\right\rangle$ is the state of object $i$, consisting of its position ${q_i}$ and velocity ${\dot{q_i}}$, and $a^o_i$ denotes physical attributes such as its radius. Each edge $r_k= \left\langle {d_k},{s_k},{a^r_k}\right\rangle$ encodes the relation between objects $i$ and $j$ with ${d_k}=q_i-q_j$ representing the displacement vector, ${s_k}=\dot q_i-\dot q_j$ denoting the velocity difference between them, and ${a^r_k}$ representing  attributes of relation $k$ such as the type of the joints connecting objects $i$ and $j$.

\paragraph*{Physics Prediction}

Propagation networks encode the states of the objects and the relations between them separately.
This encoding is carried out by two encoders, one for the relations denoted by $f^{enc}_R$ and one for the objects denoted by $f^{enc}_O$, defined as follows:
\begin{eqnarray}
    c^r_{k,t} = f^{enc}_R\left(r_{k,t}\right), \quad k=1\ldots N^r\\
    c^o_{i,t} = f^{enc}_O\left(o_{i,t}\right), \quad i=1\ldots N^o
\end{eqnarray}
where $o_{i,t}$ and $r_{k,t}$ represent the object $i$ and the relation $k$ at time $t$, respectively.

To predict the next state of the system, these encoders are used in the subsequent propagation steps within two different propagator functions, $f^{l}_R$ for relations and $f^{l}_O$ for objects, at the propagation step $l$, as follows:
\begin{eqnarray}
    e^{l}_{k,t} = f^{l}_R\left(c^r_{k,t},p^{l-1}_{i,t},p^{l-1}_{j,t}\right), \quad k=1\ldots N^r\\
    p^l_{i,t} = f^{l}_O\left (c^o_{i,t},p^{l-1}_{i,t},\sum_{k \in \mathcal{N}_i} e^{l-1}_{k,t}\right ), \quad i=1\ldots N^o
\end{eqnarray}
where $\mathcal{N}_i$ denotes the set of relations where object $i$ is being a part of, and $e^{l}_{k,t}$ and $p^l_{i,t}$ represent the propagating effects from relation $k$ and object $i$ at propagation step $l$ at time $t$, respectively. Here, the number of propagation steps can be decided depending on the complexity of the task. Through using the predicted states as inputs, it can chain the predictions and estimate the state of the objects at $t+T$. See \cite{li2018propagation} for more detailed description of this network. 

\paragraph*{Belief Regulation}

The success of physics prediction step highly depends on how accurate the environment is encoded in the graph structure. Here we refer to the term belief as the estimated world state and given previous states and motor commands, the role of the belief regulation module is to constantly update this crucial part. In propagation network \cite{li2018propagation}, the authors provide a method for estimating unknown parameters using gradient updates. In theory, it is possible to adapt this framework, however, since the relation types need to be represented as one-hot vectors, employing a continuous representation may lead to unreliable predictions. As the main theoretical contribution of this work, we propose a \emph{temporal propagation network} architecture that augments a propagation network with a recurrent neural network (RNN) unit to regulate beliefs regarding object and relation information over time. More formally, it takes a sequence of a set of state variables during the action execution as input and employing a secondary special-purpose propagation network, it encodes these structured observations, which are then fed into an RNN cell to update the current world state, as follows: 
\begin{eqnarray}
    r'_{k,t} = f^{blf}_O\left(e^{L}_{k,t},r'_{k,t-1}\right), \quad k=1\ldots N^r\\
    o'_{i,t} = f^{blf}_R\left(p^{L}_{i,t},o'_{i,t-1}\right), \quad i=1\ldots N^o)
\end{eqnarray}

where $L$ denotes the propagation step, and $f^{blf}_O$ and $f^{blf}_R$ denote the RNN-based encoder functions for objects and relations, respectively. Feeding these functions with the sequence of encoding vectors $r'_{k,t-1}$ and $o'_{i,t-1}$ allows the temporal propagation network to consider the overall history of object and relation states from the previous time-steps. Hence, it continuously updates its belief regarding objects and relations states ($o_{i,t}$ and $r_{k,t}$), and eventually minimize the difference between the effect predicted by our physics prediction module and reality.

\section{Experimental Setup}
We evaluate our model in simulation and on a real robot through a set of experiments. In the following, we explain the details of the experimental setups designed to assess the generalization performance to the changing number of objects and time steps, transferability of our model to different object-relation distributions and the real-world setting.

\subsection{Robotic Setup}
\begin{figure}[!t]
    \centering
    \begin{subfigure}[b]{0.21\linewidth}
        \centering

    \includegraphics[width=\linewidth,trim={0cm 0cm 0cm 2cm},clip]{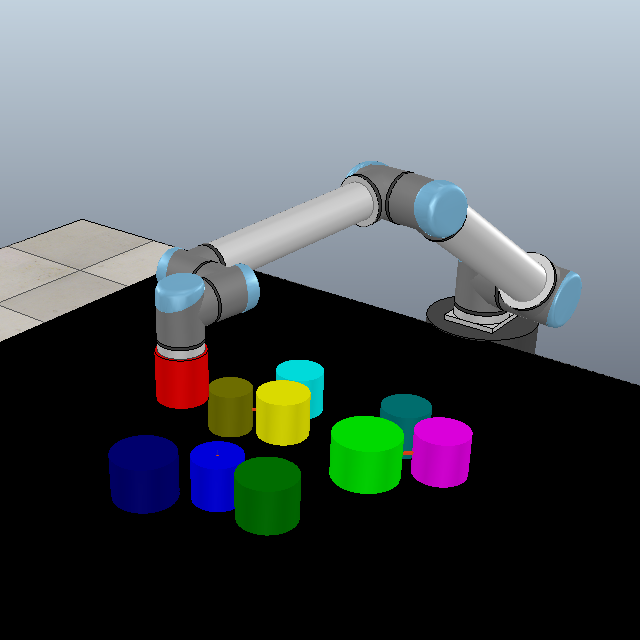}
    \caption{Simulation}
    \end{subfigure}
    \begin{subfigure}[b]{0.256\linewidth}
        \centering

    \includegraphics[width=\linewidth]{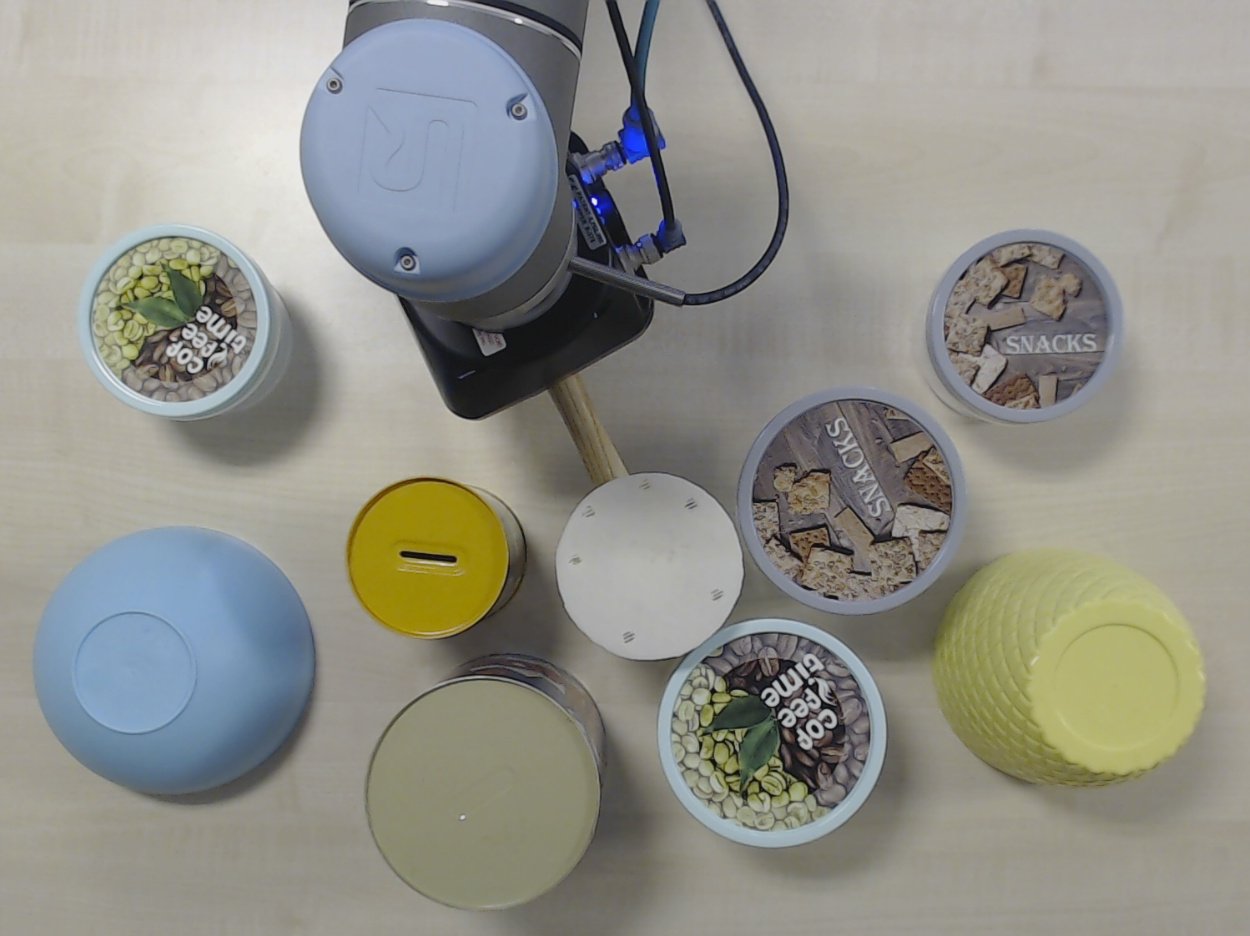}
    \caption{Real-World}
    \end{subfigure}
    \begin{subfigure}[b]{0.192\linewidth}
    \centering
    \includegraphics[width=\linewidth]{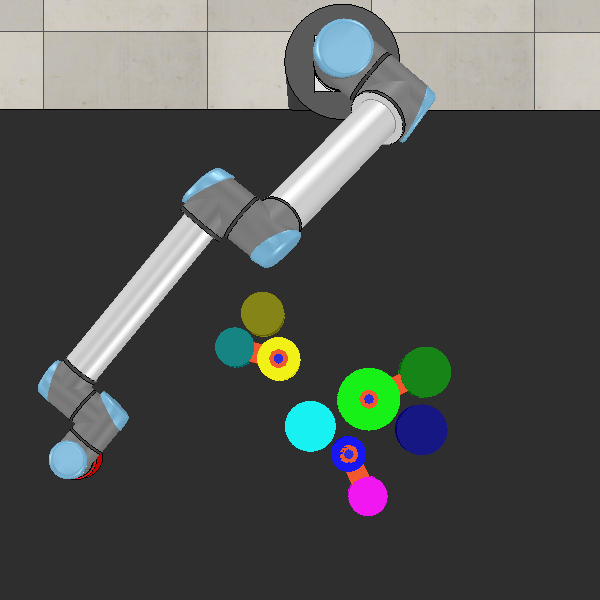}
    \caption{Sparse}\label{fig:sparse}
    \end{subfigure}
    \begin{subfigure}[b]{0.21\linewidth}
    \centering
    \includegraphics[width=\linewidth]{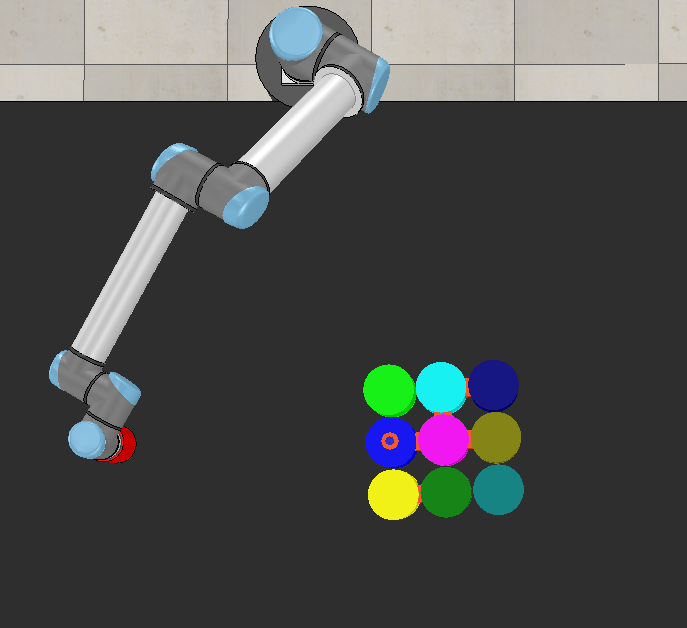}
    \caption{Dense}
    \label{fig:dense}
    \end{subfigure}
    
    \caption{Sample setups (a-b), and initial configurations (c-d)}\vspace{-0.4cm}
    \label{fig:Scenes}
\end{figure}

Our simulation and real world experiments included a 6~degrees of freedom UR10 arm and several cylindrical objects placed on a table as shown in Fig~\ref{fig:Scenes}a-b. The table-top settings were composed of objects of varying numbers and sizes. The objects might move independent of each other (no-joint) or connected through one of three different types of joints, namely \emph{fixed}, \emph{revolute} and \emph{prismatic}. We prepared two sets of environments: fixed environments, which could contain only fixed joints, and mixed environments that may contain all three joint types. The robot learned to predict the effects of its actions by self-exploration and observation in the V-REP physics-based simulator with Bullet engine\footnote{www.coppeliarobotics.com/}. The simulated robot exercised its push action on a set of objects by moving a cylindrical object that was attached to its end-effector. After training, the performance of the prediction model was tested both in the simulated and real-world settings. 

In our simulation experiments, we considered two different configurations for scene generation: a \emph{sparse} configuration (Fig.\ref{fig:sparse}) where objects were initially scattered randomly in the scene, and a \emph{dense} one (Fig.\ref{fig:dense}) where the objects were initially grouped. Sparse configuration was specifically designed to maximize the contact time between the end-effector and the objects and to allow a rich set of interactions. The robot chooses 8 different linear motions of 30cm, maximizing contact time with the most diverse set of objects. While in the sparse configuration the objects were randomly scattered in the scene, in the dense configuration the objects were grouped in a grid structure. 
Sparse configuration was used for training, and dense configuration for testing the generalization performance of the model on novel environments, i.e. on instances drawn from a completely different distribution of objects and relations. A total of 900 different 9-object scenes were used for training the model. For testing, both sparse and dense configurations are used. The sparse test set was composed of 50 9-object, 25 6-object, and 25 12-object scenes. The dense test set was composed of 50 6-object, 50 8-object, and 50 9-object scenes. For each scene above, the robot arm approached from four different random directions. Each object in these scenes had radii between 8 cm to 16 cm.
In the evaluations, separate models were trained and tested on scenes where only fixed joints and mixed types of joints exist.

\subsection{Implementation Details}
Our physic prediction module takes object position, velocity and radius as object features, and joint relation type between objects as relation features. More specifically, object encoder is a MLP with 3 hidden layers of 150 neurons, and it takes object radius and velocity as inputs. The relation encoder is a MLP with 1 hidden layer of 100 neurons. It takes radius, joint relation type, and position and velocity differences between the objects as input. While our relation propagator is an MLP with 2 hidden layers of 150 neurons, our object propagator is an MLP with one hidden layer of 100 neurons. During training, at each epoch, we validated our physics prediction module on the validation set containing instances from the sparse configuration and selected the model that has the lowest mean squared error (MSE) over 200 time-step trajectory roll-outs. 

Our belief regulation module uses the sequence of positions, velocities and radii of the objects, and predicts joint types between each pair of objects. The output of the relation propagator is connected to an LSTM with 100 hidden neurons. This LSTM is then connected to a fully connected layer to predict joint type between objects. This network was trained using sequences with 100 time steps. During training, we optimized this network with the loss coming from the predicted joint types between the time steps 50 and 100 to make sure that our model can generalize to the changing number of time steps, while not over-fitting to the position information coming from the single time steps.

For training our networks, we used batch size of 64, learning rate of 1e-3, and Adam optimizer\cite{kingma2014adam}. The learning rate is decayed by 0.8 when there was no decrease in validation loss for 20 steps. Networks are trained for 500 epochs. For physics prediction module, at each epoch, half of the randomly shuffled data is used, and for belief regulation module, each epoch contained 100 batch of trajectories randomly sampled from training data. It took about one day to train the physic prediction module and one day to train belief regulation module using GTX 1080Ti GPU.

\section{Results}

For quantitative analysis, we compared our method with PropNets with alternative (hard-coded) relation assignments: As a strong baseline, \textbf{PropNet$_{gt}$} uses ground-truth relations. \textbf{PropNet$_{f}$} assumes all pairs of contacting objects have fixed relations between them. \textbf{PropNet$_{n}$} assumes no joints between objects. Furthermore, to analyze the influence of temporal data in predicting relations within our model, we also report results with \textbf{1-step BRDPN} that predicts object relations using only the observation from the previous step.

\subsection{Quantitative Analysis of Separate Modules in Simulation}

\begin{figure}[t!]
    \centering
    \begin{subfigure}[b]{0.48\linewidth}
    \includegraphics[width=\linewidth,height=0.5\linewidth]{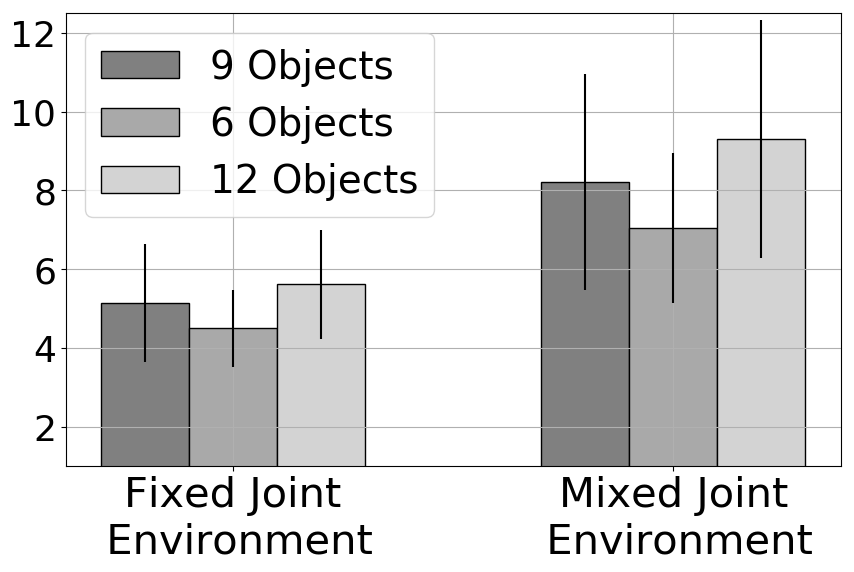}
    \caption{Sparse configuration} \label{fig:MSE-Setup1}
    \end{subfigure}
    \begin{subfigure}[b]{0.48\linewidth}
    \includegraphics[width=\linewidth,height=0.5\linewidth]{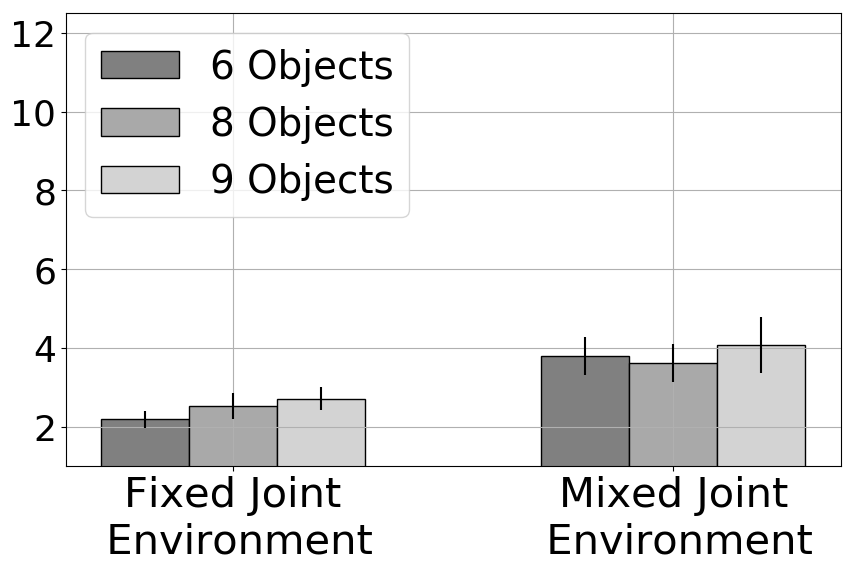}
    \caption{Dense configuration}
    \end{subfigure}

    \caption{Error(cm) for object positions over (a) 200 time-step trajectory roll-outs for sparse configuration, and (b) 50 time-step trajectory roll-outs for dense configuration.}\vspace{-0.5cm}
    \label{fig:MSE-PROP}
\end{figure}

First, the physics prediction module is evaluated given ground-truth relation information.
Fig.~\ref{fig:MSE-PROP} presents the performance on the test set for different object configurations. Each bar provides the mean error averaged over differences between predicted and observed trajectories. We evaluated for both the sparse and dense configuration settings in {\em fixed-joint} and {\em mixed-joint} environments separately. As shown, around $7$ and $3$ cm mean error is observed in sparse and dense object configurations. Furthermore, we observed that the error drops significantly (to $4$ and $2$ cm) in case only fixed joints are included. Given the average motion (including zero motion in many cases) in objects is $40 cm$ sparse and $18 cm$ in dense configuration, these results show that the model achieves high prediction performance if it uses the ground-truth relations; and with the increasing complexity of object relations, learning becomes more challenging.

\begin{figure}[!t]
    \centering
    \includegraphics[width=\linewidth]{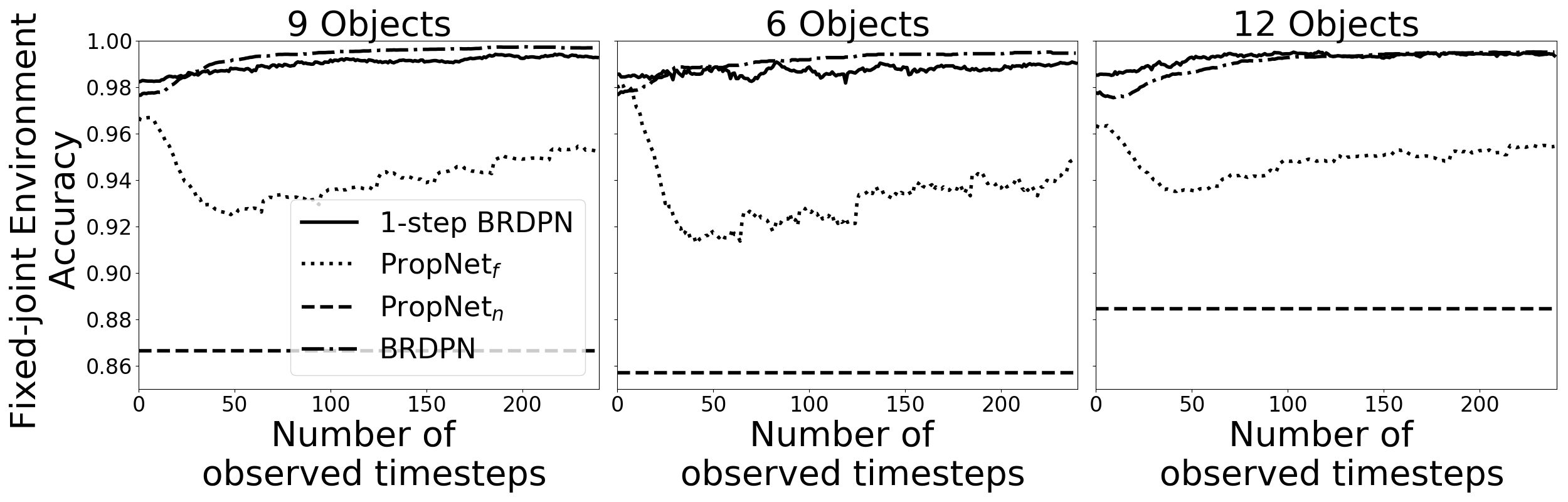}
    
    \includegraphics[width=\linewidth]{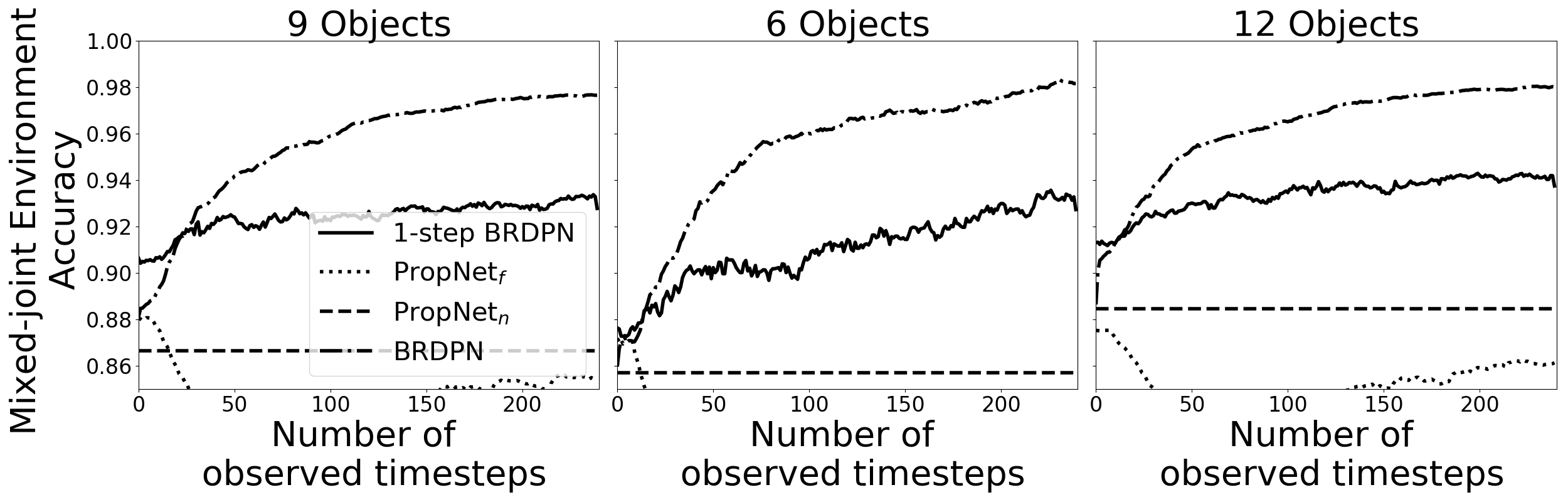}
    \caption {Relation prediction accuracies (sparse configuration).}
    \label{fig:Accuracy-PROP}
\end{figure}
\begin{figure}[h]
    \centering
\setlength{\fboxsep}{0pt}

 \begin{subfigure}[b]{0.30\linewidth}
     \centering
    \fbox{\includegraphics[width=\linewidth,trim={1cm 2cm 0.5cm 2cm},clip]{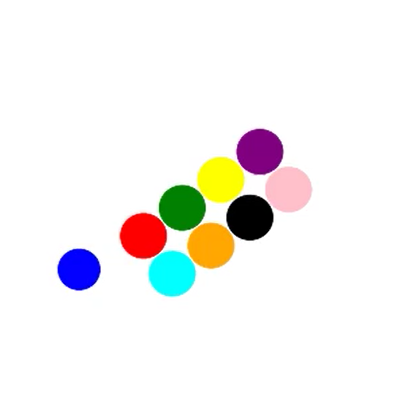}}
    \caption{Initial Scene} 
    \end{subfigure}
\begin{subfigure}[b]{0.30\linewidth}
    \centering
    \fbox{\includegraphics[width=\linewidth,trim={1cm 2cm 0.5cm 2cm},clip]{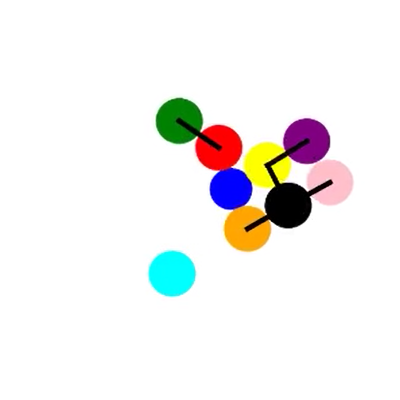}}
    \caption{\centering Predicted}
\end{subfigure}
 \begin{subfigure}[b]{0.30\linewidth}
    \centering
    \fbox{\includegraphics[width=\linewidth,trim={1cm 2cm 0.5cm 2cm},clip]{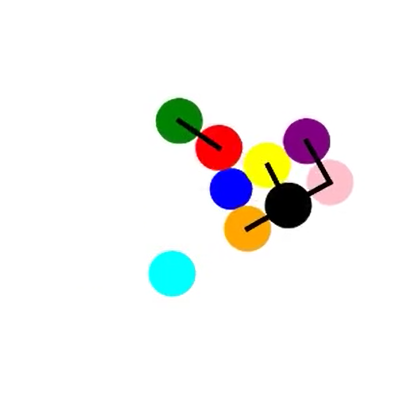}}
    \caption{\centering Ground-Truth}
\end{subfigure}

    \caption{ (a) the setting is shown, where the end effector of the robot (shown in blue) moves towards an object group. (b) figure shows the predictions on fixed joint relations (black lines) and (c) provides the ground-truth relations.}\vspace{-0.5cm}
    \label{fig:predicted_relations}
\end{figure}

Next, the performance of our belief regulation prediction module is evaluated on the sparse test set. As shown in Fig.~\ref{fig:Accuracy-PROP}, the accuracy is already very high from the instant when the robot makes its first contact in {\em fixed-joint} environments. The accuracy increases in {\em mixed-joint} environments to over 98\% as well, with the accumulated observations from the interactions of the robot.

We performed several experiments in the dense configuration as well. However, we observed that directly comparing real and predicted relations in this configuration might be misleading as different sets of joints that connect the objects in the same grid might generate identical effects in response to the robot interactions. The system might suffer from ambiguities in predicting joint relations from such interaction experience. For example, a group of objects that form a rigid body through a different set of connections would behave the same in response to the push action. Fig.~\ref{fig:predicted_relations} provides a snapshot of such a case where the robot started interaction with 8 objects placed on a grid. 
In this case, even if the joint relations were incorrectly predicted for the sub-group of 5 objects, this was a plausible inference that enabled the system to make correct predictions about the object trajectories from that moment. While incorrect state predictions might not affect the effect prediction performance of the system in this particular extreme example, we might need intelligent exploration strategies that enable the robot to collect more reliable information in other ambiguous cases.

\begin{figure}[t!]
    \centering
    \captionsetup{justification=centering}
    
    \includegraphics[width=\linewidth]{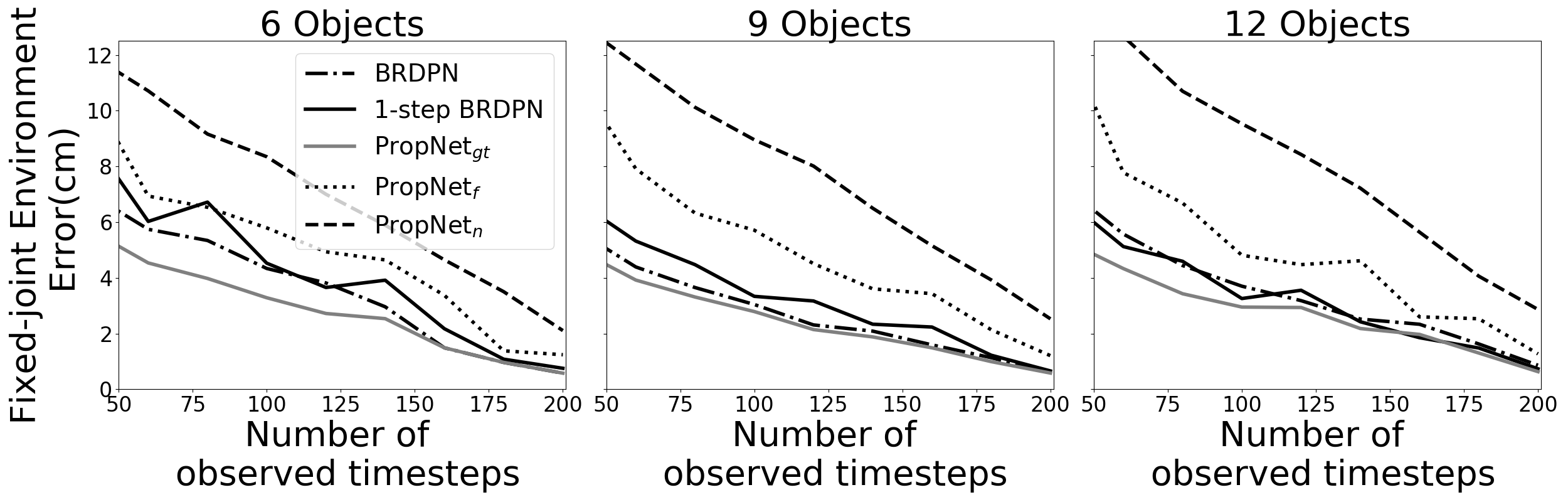}

    \includegraphics[width=\linewidth]{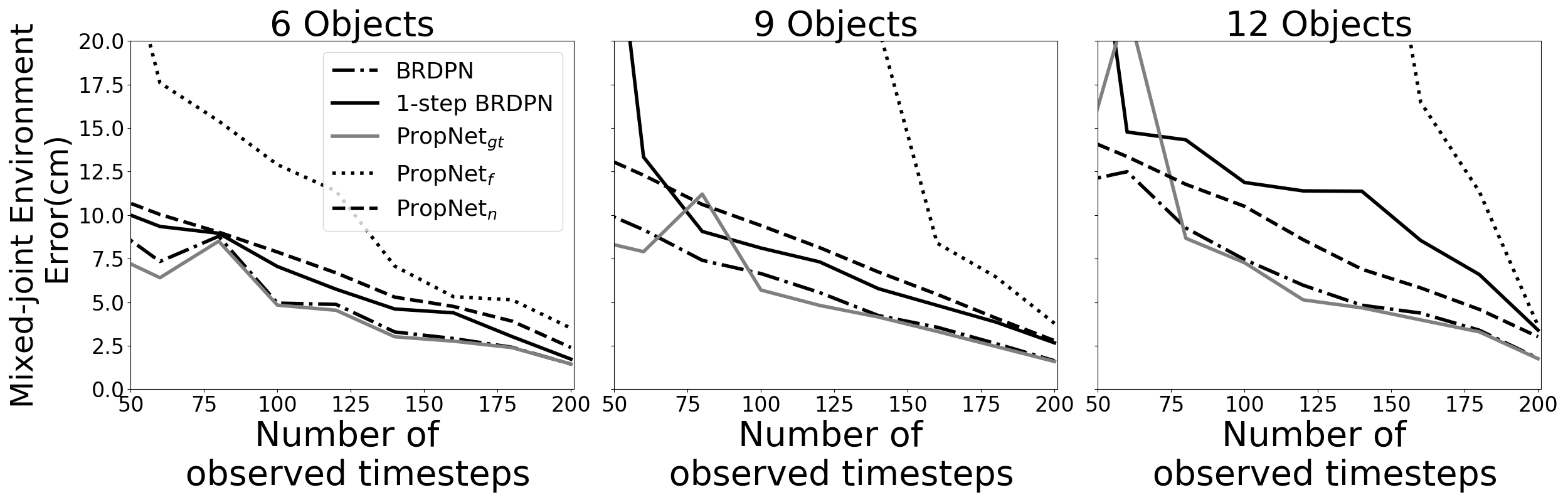}
    
    \caption {Error of the BRDPN in sparse configuration.}
    \label{fig:BR-Sparse}
\end{figure}
\begin{figure}[h]
    \centering
    \captionsetup{justification=centering}
    
    \includegraphics[width=\linewidth]{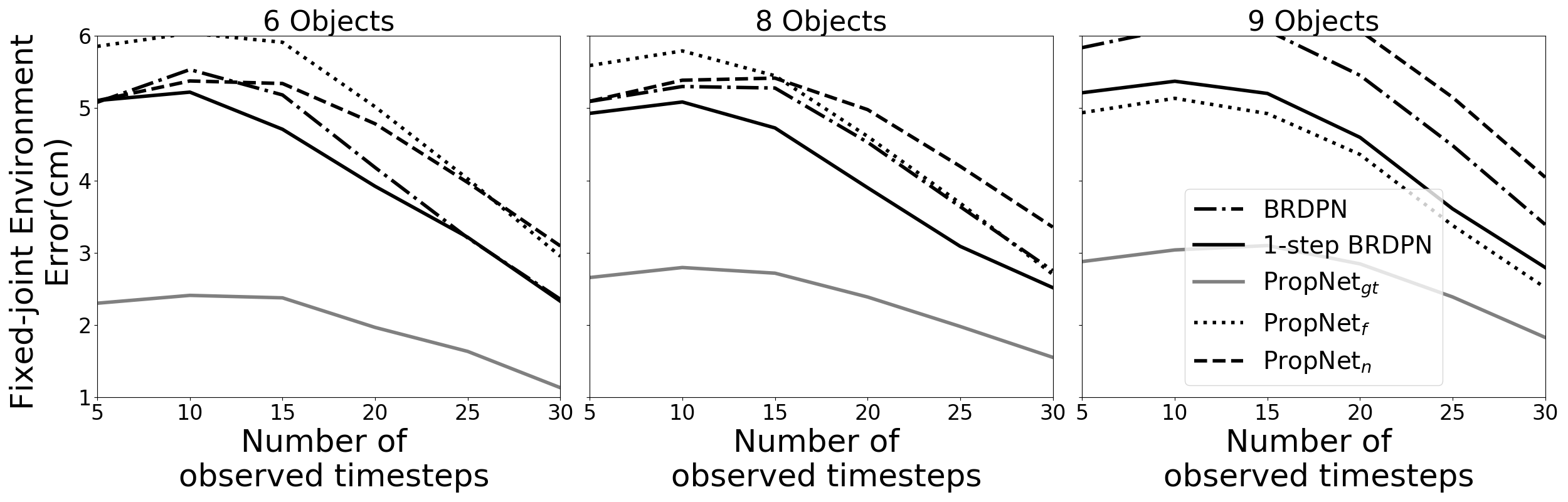}

    \includegraphics[width=\linewidth]{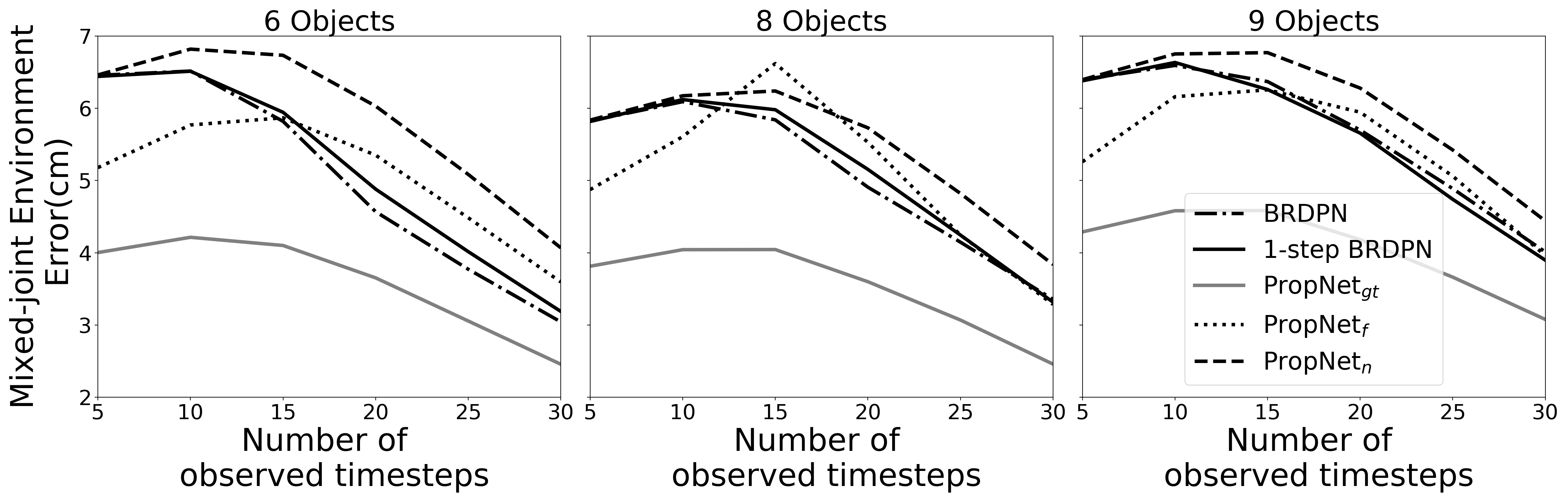}
    
    \caption {Error of the BRDPN in dense configuration.}\vspace{-0.5cm}
    \label{fig:BR-Dense}
\end{figure}

\subsection{Quantitative Analysis of the BRDPN in Simulation} \label{sec:epwpsi}

In this section, the complete system is evaluated at different time-points during interactions. The belief regulation module predicts the relations between objects using the observations up to the corresponding time-points. Given the states of the objects, the robot actions, and the predicted relations between pairs of objects, the physics prediction module finds the trajectories of the objects that are expected to be observed for the rest of the motion. The results are presented in Fig.~\ref{fig:BR-Sparse} and Fig.~\ref{fig:BR-Dense} where the errors on the remaining trajectories are computed with the predictions of the system at the reported time steps. These results indicate that even if the relations are unknown, the proposed belief regulation improves the effect prediction performance of the system with more interaction experience. While BRDPN performs better for the sparse dataset, 1-step BRDPN performs similar to or better than BRDPN in dense configurations probably because the model was optimized for temporal information coming from sparse environments. Note that BRDPN outperforms the PropNets variants that do not utilize the ground-truth relations. PropNet$_{gt}$ has the best performance since it has access to the ground-truth relation assignments. 

\setlength{\fboxsep}{0pt}

\begin{figure}[!t]
    \centering
    \raisebox{0.18in}{\rotatebox[origin=t]{90}{Demo}}
    \includegraphics[width=.35\linewidth]{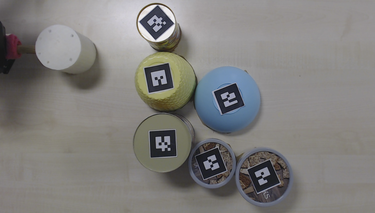}
    \hspace{-0.25cm}
    \includegraphics[width=.35\linewidth]{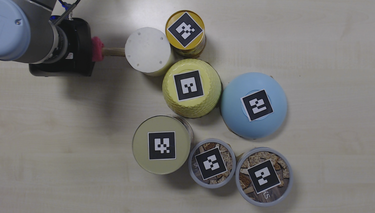}
    
    \hspace{0.3cm}
     $t: 10 \xrightarrow{\hspace{0.2\linewidth}} t: 60$\vspace{-0.35cm}

    \raisebox{-0.4in}{\rotatebox[origin=t]{90}{\parbox{1.5 cm}{PropNet$_f$}}}
    \fbox{\includegraphics[height=0.35\linewidth,width=0.2\linewidth,angle=-90,trim={5cm 2cm 3cm 3cm},clip]{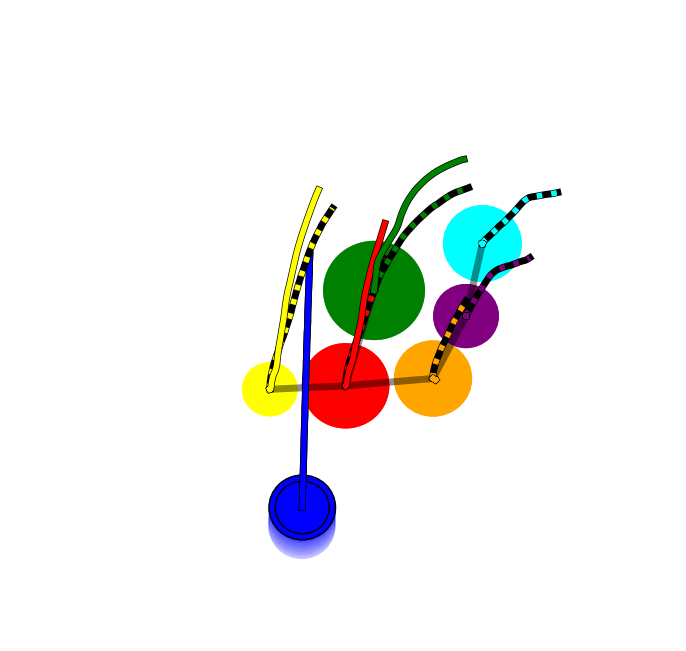}}
    \hspace{-0.25cm}
    \fbox{\includegraphics[height=0.35\linewidth,width=0.2\linewidth,angle=-90,trim={5cm 2cm 3cm 3cm},clip]{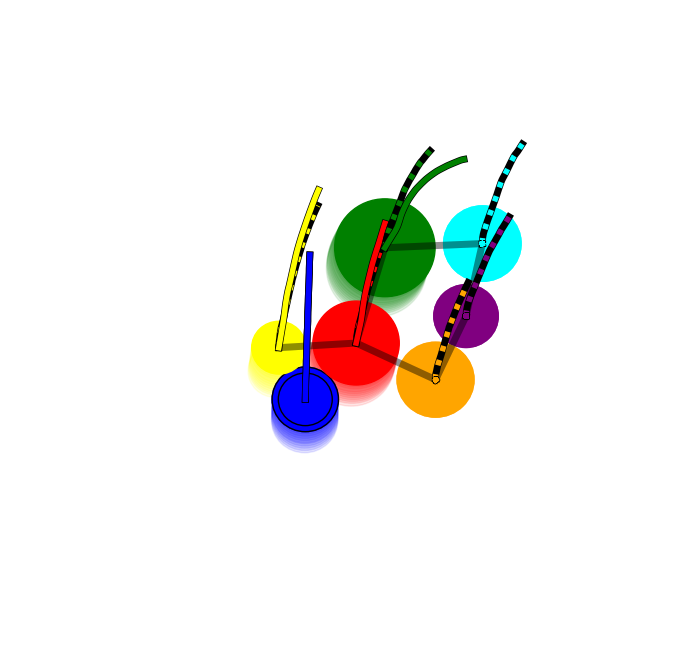}}

    \hspace{0.02 cm}\raisebox{-0.4in}{\rotatebox[origin=t]{90}{\parbox{1.5 cm}{PropNet$_n$}}}
    \fbox{\includegraphics[height=0.35\linewidth,width=0.2\linewidth,angle=-90,trim={5cm 2cm 3cm 3cm},clip]{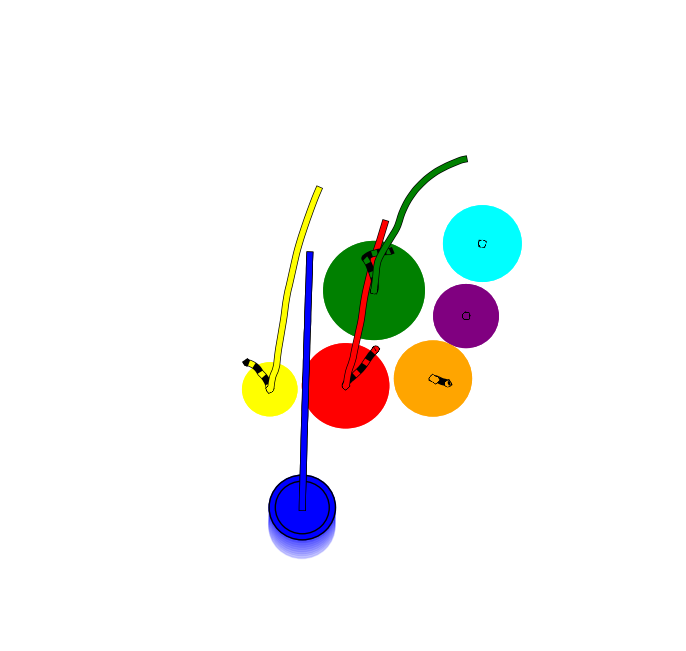}}
    \hspace{-0.25cm}
    \fbox{\includegraphics[height=0.35\linewidth,width=0.2\linewidth,angle=-90,trim={5cm 2cm 3cm 3cm},clip]{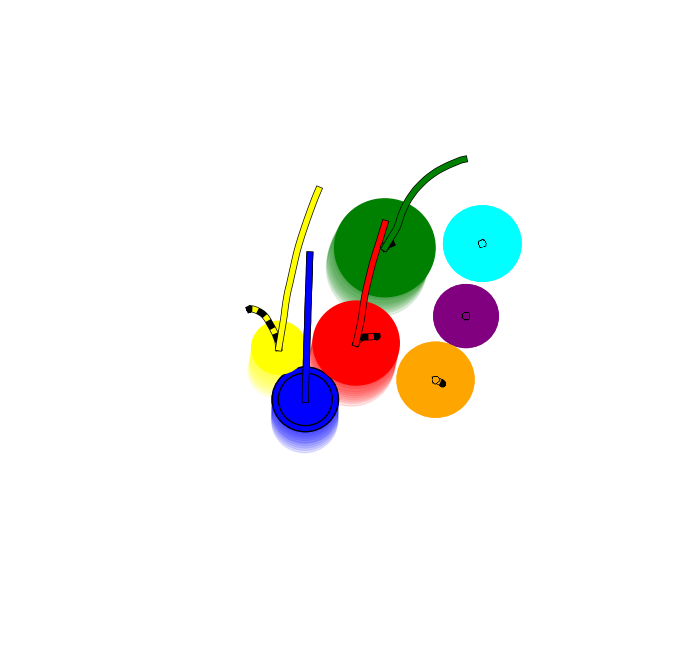}}
    
    \hspace{0.1 cm}\raisebox{-0.4in}{\rotatebox[origin=t]{90}{\parbox{1.5 cm}{BRDPN}}}
    \fbox{\includegraphics[height=0.35\linewidth,width=0.2\linewidth,angle=-90,trim={5cm 2cm 3cm 3cm},clip]{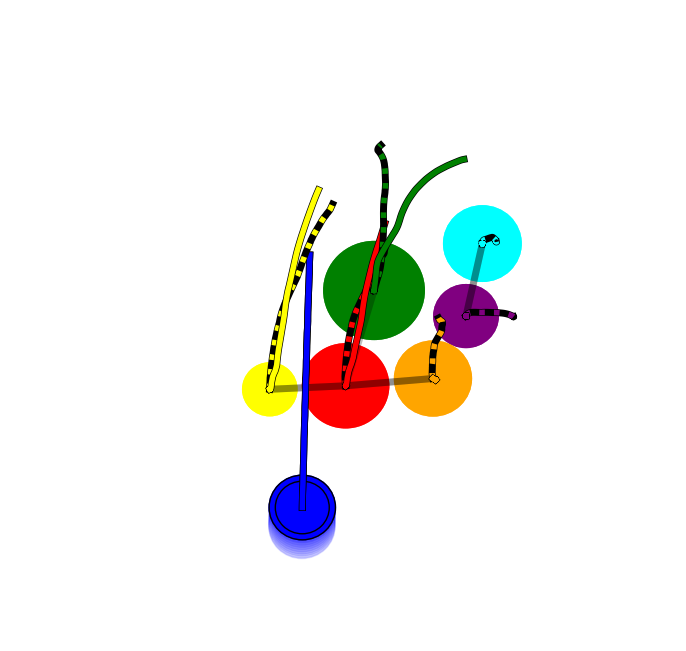}}
    \hspace{-0.25cm}
    \fbox{\includegraphics[height=0.35\linewidth,width=0.2\linewidth,angle=-90,trim={5cm 2cm 3cm 3cm},clip]{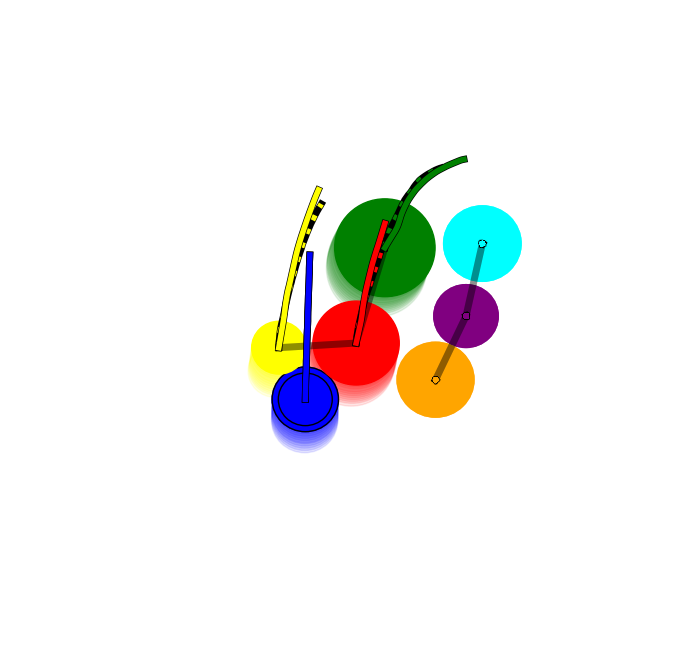}}

    \hspace{0.02 cm}\raisebox{-0.4in}{\rotatebox[origin=t]{90}{\parbox{1.5 cm}{PropNet$_{gt}$}}}
    \fbox{\includegraphics[height=0.35\linewidth,width=0.2\linewidth,angle=-90,trim={5cm 2cm 3cm 3cm},clip]{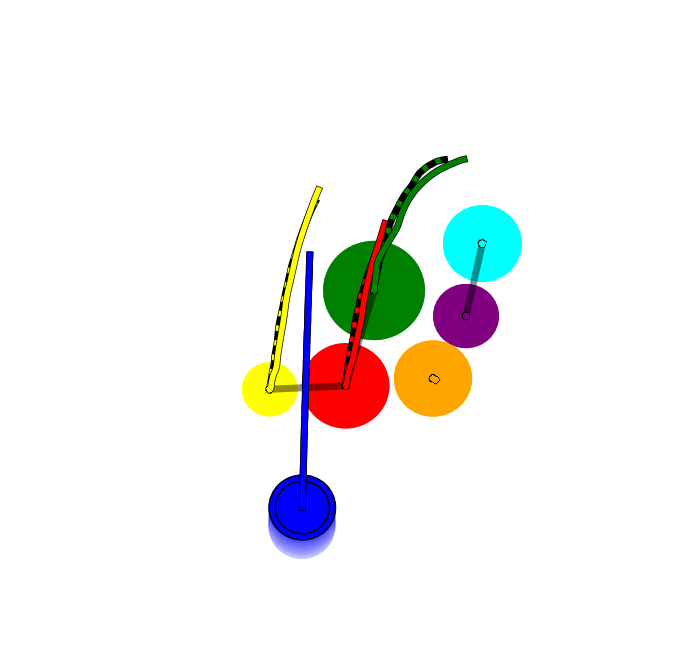}}
    \hspace{-0.25cm}
    \fbox{\includegraphics[height=0.35\linewidth,width=0.2\linewidth,angle=-90,trim={5cm 2cm 3cm 3cm},clip]{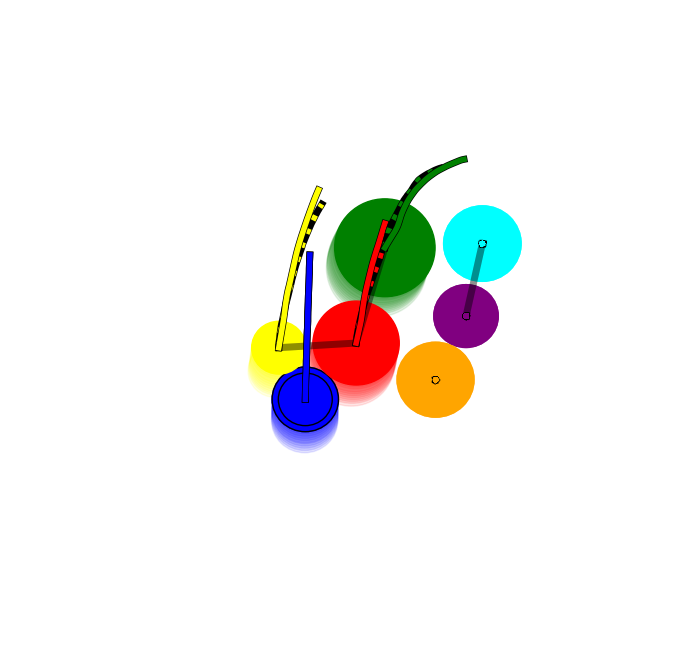}}
    
    \caption {The first real-world interaction example. The relation assignments/predictions (black), the real (solid colored) and the predicted (dashed colored) trajectories are shown.}
    \label{fig:real_world_pred1}\vspace{-0.3cm}
\end{figure}

\subsection{Real World Experiments}

In this section, we provide the results obtained in the real world. For this, the prediction model trained in the simulator was directly transferred to the real world. A mallet that was grasped by the 3-finger gripper of the UR10 robot was used to push objects The cylindrical objects on the setup can be seen in the Fig.~\ref{fig:Scenes}. Only one type of joint, namely fixed joint was used in this setup. Fixed joint relations are accomplished by placing customized card-boards under the specified objects, making all the group move together.
A top-down oriented RGB camera with $1920 \times 1080$ pixels resolution was placed above the scene, ARTags were placed on the objects for tracking.

\begin{figure}[t!]
    \centering
     \includegraphics[width=0.12\textwidth]{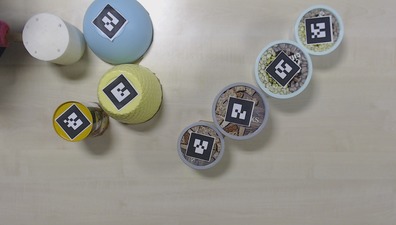}
    \hspace{-0.25cm}
    \includegraphics[width=0.12\textwidth]{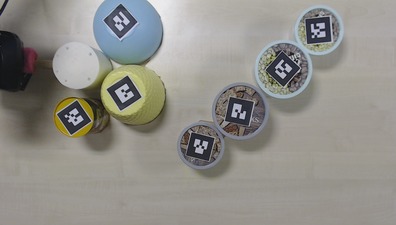}
    \hspace{-0.245cm}
    \includegraphics[width=0.12\textwidth]{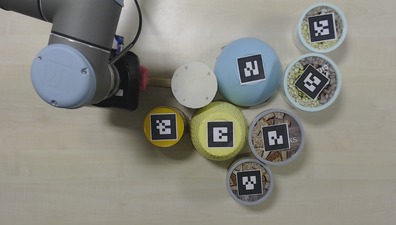}
    \hspace{-0.245cm}
    \includegraphics[width=0.12\textwidth]{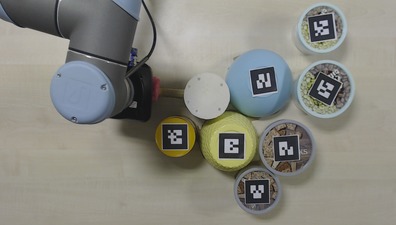}
    \vspace{-.16cm}
    
    \hspace{0.02\textwidth}
    $t: 0 \xrightarrow{\hspace{0.045\textwidth}} 
    t: 50 \xrightarrow{\hspace{0.045\textwidth}} t: 140 \xrightarrow{\hspace{0.045\textwidth}}
    t: 150$\hspace{0.02\textwidth}
    \vspace{-.26cm}
    
    \fbox{\includegraphics[height=0.12\textwidth,angle=-90,trim={2cm 0 5cm 0},clip]{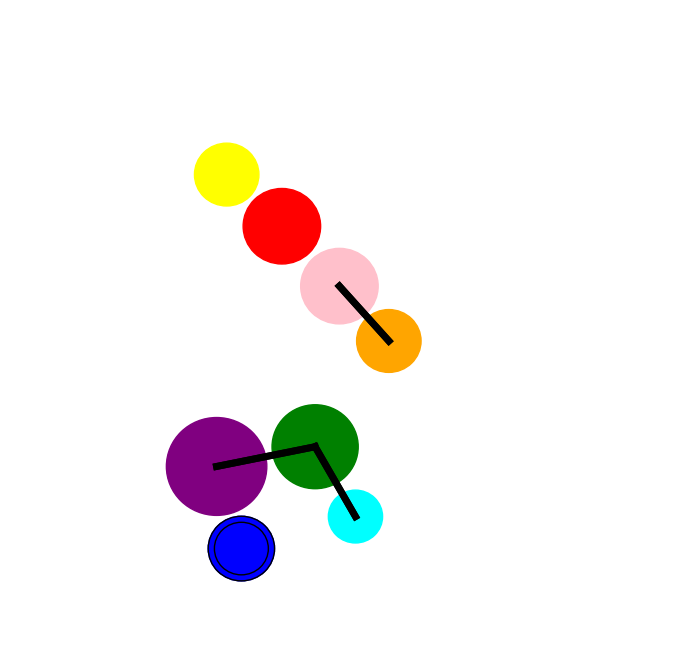}}
    \hspace{-0.25cm}
    \fbox{\includegraphics[height=0.12\textwidth,angle=-90,trim={2cm 0 5cm 0},clip]{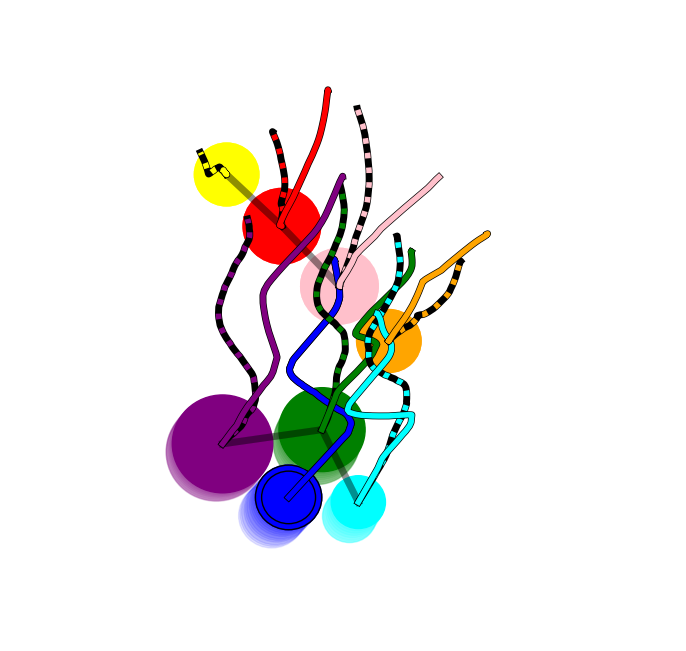}}
    \hspace{-0.25cm}
    \fbox{\includegraphics[height=0.12\textwidth,angle=-90,trim={2cm 0 5cm 0},clip]{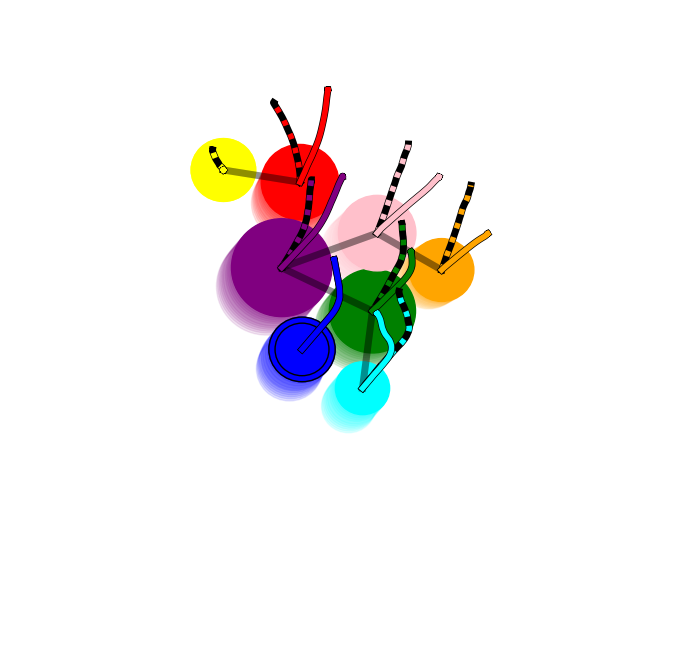}}
    \hspace{-0.25cm}
    \fbox{\includegraphics[height=0.12\textwidth,angle=-90,trim={2cm 0 5cm 0},clip]{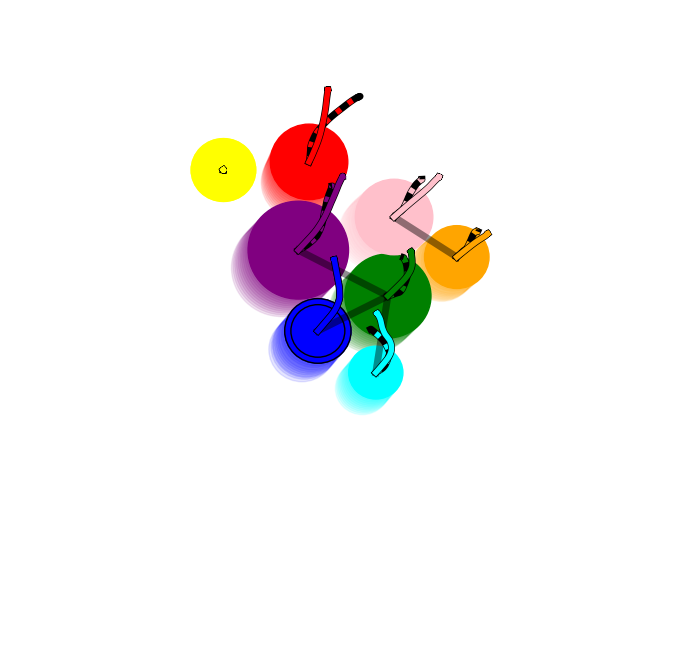}}
    \caption {
     The second real-world interaction example. } \vspace{-0.5cm}
    \label{fig:real_world_pred2}
\end{figure}

First, we present the results qualitatively over two example scenes. In the first example scene, 6 objects were placed as a group as shown in Fig.~\ref{fig:real_world_pred1}, where the top left 3 objects and the bottom right pair were connected to each other. A straight push motion was executed by the robot and the object positions at time steps 10 and 60 were provided. Solid and dashed lines show the real and predicted trajectories. As shown, given ground-truth joint information, the model made almost perfect trajectory predictions. When the ground-truth relations are not provided, as in PropNet$_{n}$ and PropNet$_{f}$, the model either predict that all objects are pushed aside or all contacting ones move together. Finally, when the relations are predicted, first the model predicts trajectories similar to PropNet$_{f}$ case, but after seeing the independent motions of upper three object group, it corrects the joint relations and predicts the correct trajectory successfully. In the second example scene, a more challenging configuration was used, where 7 objects were placed in two separate groups and objects in each group are attached to each other (Fig.~\ref{fig:real_world_pred2}). The end-effector made a zigzag motion towards the objects. The relation prediction on the first group (the one closer to the robot) was correct at $t: 50$, since the robot had sufficient interaction with these objects. The indirect contact to the second group via the first group took place slightly before  $t: 140$ and the robot correctly inferred that not all the objects in the second group had fixed relations. The prediction of the first group remained correct, but the robot made incorrect predictions in two cases: it incorrectly inferred that the first and second group was connected and that the top-right pair was also connected. With further interaction, these incorrect inferences were corrected at $t:150$. 

\begin{figure}[!t]
    \centering
    \includegraphics[width=\linewidth,height=0.18\linewidth]{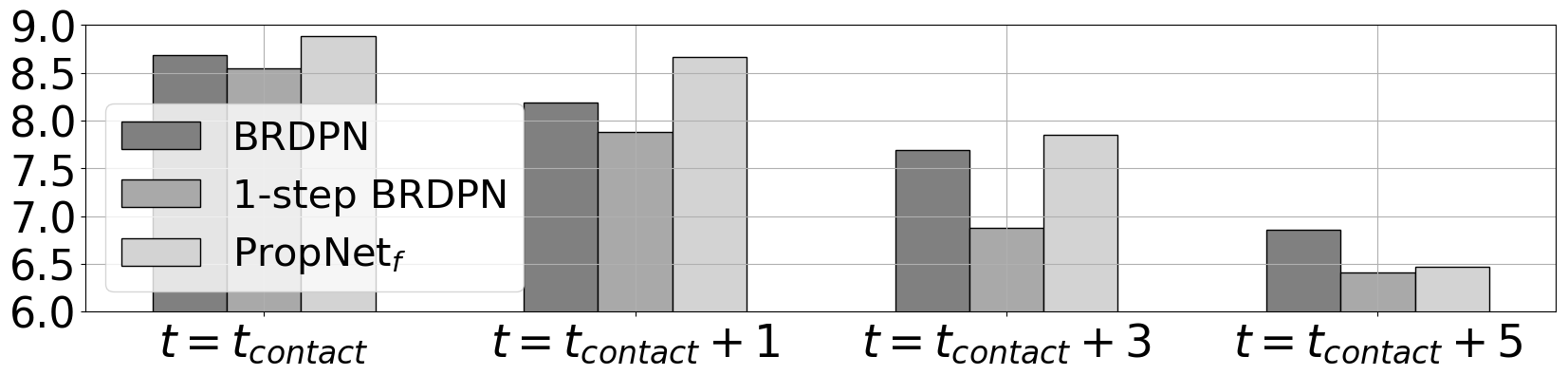}
    
    \caption {Average error (in cm) in the real world.}
    \label{fig:MSE-BRDPN-REAL} \vspace{-0.6cm}
\end{figure}

Finally, we evaluate our model quantitatively with a large number of interactions. We generated 102 different setups that include 2 to 5 objects with 1 to 3 connections. One of the 5 different predefined straight motions of 30 to 60~cm was applied towards these objects that were placed in different locations which results in objects moving $19.5$~cm on average. Our model achieved an average error of $6.6$~cm in predicting their final positions. Although in some cases incorrect effect predictions caused failures in predicting the movement direction of interacted objects, our model performed well considering the average diameter of $12$~cm of the objects and our direct transfer strategy from the simulation. Fig.~\ref{fig:MSE-BRDPN-REAL} provides a more detailed analysis of the results focusing on the time-point when the first contact with the objects occur. As shown, the prediction error of 1-step BRDPN quickly drops compared to the model that assigns fixed-joint to objects whose distances are smaller than $2.5$~cm. Probably after the objects physically separated from each other, PropNet$_{f}$ does not consider those objects to be attached to each other and also start making predictions with similar accuracy. Note that PropNet$_{n}$ significantly under-performed and was not included in the figure, and ground-truth-relation model generated higher performance consistently, obtaining around $4$ cm error at the end.

\section{Conclusion}
We presented \emph{Belief Regulated Dual Propagation Networks (BRDPN)}, a general-purpose learnable physics engine that also continuously updates the estimated world state through observing the consequences of its interactions. We demonstrated our network in setups containing articulated multi-part multi-objects settings. In these settings, we validated our network and its modules on several test cases 
which have shown both strengths and weaknesses of the proposed methods. As future work, we aim to include new experiments that provide more detailed comparisons of proposed methods. While our system was validated in both simulation and real-world robotic experiments, we discussed that intelligent exploration strategies that resolve the inference problem in ambiguous situations are necessary. In the future, we aim to study on generating goal-directed action trajectories that balance the trade-off between exploration and exploitation. Furthermore, we plan to use the learned effect predictions in making multi-step plans in potentially sub-symbolic \cite{ugur2011goal} or symbolic \cite{ugur2015bottom} spaces.

%\clearpage
\addtolength{\textheight}{-11.5cm}
\bibliographystyle{IEEEtran}
\bibliography{references} 

\begin{thebibliography}{10}
\providecommand{\url}[1]{#1}
\csname url@rmstyle\endcsname
\providecommand{\newblock}{\relax}
\providecommand{\bibinfo}[2]{#2}
\providecommand\BIBentrySTDinterwordspacing{\spaceskip=0pt\relax}
\providecommand\BIBentryALTinterwordstretchfactor{4}
\providecommand\BIBentryALTinterwordspacing{\spaceskip=\fontdimen2\font plus
\BIBentryALTinterwordstretchfactor\fontdimen3\font minus
  \fontdimen4\font\relax}
\providecommand\BIBforeignlanguage[2]{{%
\expandafter\ifx\csname l@#1\endcsname\relax
\typeout{** WARNING: IEEEtran.bst: No hyphenation pattern has been}%
\typeout{** loaded for the language `#1'. Using the pattern for}%
\typeout{** the default language instead.}%
\else
\language=\csname l@#1\endcsname
\fi
#2}}

\bibitem{SEKER2019173}
\BIBentryALTinterwordspacing
M.~Y. Seker, A.~E. Tekden, and E.~Ugur, ``Deep effect trajectory prediction in
  robot manipulation,'' \emph{Robotics and Autonomous Systems}, vol. 119, pp.
  173 -- 184, 2019. [Online]. Available:
  \url{http://www.sciencedirect.com/science/article/pii/S0921889019300740}
\BIBentrySTDinterwordspacing

\bibitem{battaglia2016interaction}
P.~Battaglia, R.~Pascanu, M.~Lai, D.~J. Rezende, \emph{et~al.}, ``Interaction
  networks for learning about objects, relations and physics,'' in
  \emph{Advances in neural information processing systems}, 2016, pp.
  4502--4510.

\bibitem{chang2016compositional}
M.~B. Chang, T.~Ullman, A.~Torralba, and J.~B. Tenenbaum, ``A compositional
  object-based approach to learning physical dynamics,'' in \emph{International
  Conference on Learning Representations}, 2017.

\bibitem{li2018propagation}
Y.~Li, J.~Wu, J.-Y. Zhu, J.~B. Tenenbaum, A.~Torralba, and R.~Tedrake,
  ``Propagation networks for model-based control under partial observation,''
  in \emph{International Conference on Robotics and Automation}, 2019.

\bibitem{mrowca2018flexible}
D.~Mrowca, C.~Zhuang, E.~Wang, N.~Haber, L.~F. Fei-Fei, J.~Tenenbaum, and D.~L.
  Yamins, ``Flexible neural representation for physics prediction,'' in
  \emph{Advances in neural information processing systems}, 2018, pp.
  8813--8824.

\bibitem{li2018learning}
Y.~Li, J.~Wu, R.~Tedrake, J.~B. Tenenbaum, and A.~Torralba, ``Learning particle
  dynamics for manipulating rigid bodies, deformable objects, and fluids,'' in
  \emph{International Conference on Learning Representations}, 2019.

\bibitem{watters2017visual}
N.~Watters, D.~Zoran, T.~Weber, P.~Battaglia, R.~Pascanu, and A.~Tacchetti,
  ``Visual interaction networks: Learning a physics simulator from video,'' in
  \emph{Advances in neural information processing systems}, 2017, pp.
  4539--4547.

\bibitem{van2018relational}
S.~van Steenkiste, M.~Chang, K.~Greff, and J.~Schmidhuber, ``Relational neural
  expectation maximization: Unsupervised discovery of objects and their
  interactions,'' in \emph{International Conference on Learning
  Representations}, 2018.

\bibitem{battaglia2013simulation}
P.~W. Battaglia, J.~B. Hamrick, and J.~B. Tenenbaum, ``Simulation as an engine
  of physical scene understanding,'' \emph{Proceedings of the National Academy
  of Sciences}, vol. 110, no.~45, pp. 18\,327--18\,332, 2013.

\bibitem{deisenroth2011pilco}
M.~Deisenroth and C.~E. Rasmussen, ``Pilco: A model-based and data-efficient
  approach to policy search,'' in \emph{Proceedings of the 28th International
  Conference on machine learning (International Conference on Machine
  Learning)}, 2011, pp. 465--472.

\bibitem{wu2015galileo}
J.~Wu, I.~Yildirim, J.~J. Lim, B.~Freeman, and J.~Tenenbaum, ``Galileo:
  Perceiving physical object properties by integrating a physics engine with
  deep learning,'' in \emph{Advances in neural information processing systems},
  2015, pp. 127--135.

\bibitem{lerer2016learning}
A.~Lerer, S.~Gross, and R.~Fergus, ``Learning physical intuition of block
  towers by example,'' in \emph{International Conference on Machine Learning},
  2016.

\bibitem{mottaghi2016newtonian}
R.~Mottaghi, H.~Bagherinezhad, M.~Rastegari, and A.~Farhadi, ``Newtonian scene
  understanding: Unfolding the dynamics of objects in static images,'' in
  \emph{Proceedings of the IEEE Conference on Computer Vision and Pattern
  Recognition}, 2016, pp. 3521--3529.

\bibitem{mottaghi2016happens}
R.~Mottaghi, M.~Rastegari, A.~Gupta, and A.~Farhadi, ``"what happens if..."
  learning to predict the effect of forces in images,'' in \emph{European
  Conference on Computer Vision}.\hskip 1em plus 0.5em minus 0.4em\relax
  Springer, 2016, pp. 269--285.

\bibitem{fragkiadaki2015learning}
K.~Fragkiadaki, P.~Agrawal, S.~Levine, and J.~Malik, ``Learning visual
  predictive models of physics for playing billiards,'' in \emph{International
  Conference on Learning Representations}, 2016.

\bibitem{hochreiter1997long}
S.~Hochreiter and J.~Schmidhuber, ``Long short-term memory,'' \emph{Neural
  computation}, vol.~9, no.~8, pp. 1735--1780, 1997.

\bibitem{finn2016unsupervised}
C.~Finn, I.~Goodfellow, and S.~Levine, ``Unsupervised learning for physical
  interaction through video prediction,'' in \emph{Advances in neural
  information processing systems}, 2016, pp. 64--72.

\bibitem{xingjian2015convolutional}
S.~Xingjian, Z.~Chen, H.~Wang, D.-Y. Yeung, W.-K. Wong, and W.-c. Woo,
  ``Convolutional lstm network: A machine learning approach for precipitation
  nowcasting,'' in \emph{Advances in neural information processing systems},
  2015, pp. 802--810.

\bibitem{byravan2017se3}
A.~Byravan and D.~Fox, ``{SE3-Nets}: Learning rigid body motion using deep
  neural networks,'' in \emph{International Conference on Robotics and
  Automation}, 2017, pp. 173--180.

\bibitem{kingma2014adam}
D.~P. Kingma and J.~Ba, ``Adam: A method for stochastic optimization,''
  \emph{arXiv preprint arXiv:1412.6980}, 2014.

\bibitem{ugur2011goal}
E.~Ugur, E.~Oztop, and E.~Sahin, ``Goal emulation and planning in perceptual
  space using learned affordances,'' \emph{Robotics and Autonomous Systems},
  vol.~59, no. 7-8, pp. 580--595, 2011.

\bibitem{ugur2015bottom}
E.~Ugur and J.~Piater, ``Bottom-up learning of object categories, action
  effects and logical rules: From continuous manipulative exploration to
  symbolic planning,'' in \emph{{International Conference on Robotics and
  Automation}}.\hskip 1em plus 0.5em minus 0.4em\relax IEEE, 2015, pp.
  2627--2633.

\end{thebibliography}

\end{document}